\documentclass{article}
\usepackage{ijcai19}

\usepackage{times}
\usepackage{soul}
\usepackage{url}
\usepackage[hidelinks]{hyperref}
\usepackage[utf8]{inputenc}
\usepackage[small]{caption}
\usepackage{graphicx}
\usepackage{amsmath}
\usepackage{booktabs}
\usepackage{algorithm}
\usepackage{algorithmic}
\usepackage{graphicx,paralist,subcaption}
\newtheorem{theorem}{Theorem}
\newtheorem{proof}{Proof}

\newtheorem{definition}{Definition}
\newtheorem{lemma}[theorem]{Lemma}
\usepackage{color,amssymb,amsmath}

\usepackage{authblk}

\author[1]{Mark Wallace}
\author[1]{Aldeida Aleti}

\affil[1]{Monash University, Wellington Road, Clayton, Vic and 3800, Australia}

\begin{document}
\title{The Neighbours' Similar Fitness Property for Local Search}

\maketitle
\begin{abstract}
For most practical optimisation problems local search outperforms random sampling --  despite the ``No Free Lunch Theorem''. This paper introduces a property of search landscapes termed {\em Neighbours' Similar Fitness} (NSF) that underlies the good performance of neighbourhood search in terms of {\em local improvement}. Though necessary, NSF is not sufficient to ensure that searching for improvement among the neighbours of a good solution is better than random search. The paper introduces an additional (natural) property which supports a general proof that, for NSF landscapes, neighbourhood search beats random search.

\end{abstract}

\section{Introduction}

Local Search is a successful class of methods used to solve many
large complex optimisation problems.
A problem $(S,f)$ is defined as a set $S$ of candidate solutions, termed its {\em search space},
and a fitness function $f$ that maps candidate solutions to a fitness measure.

Many researchers have explored why different forms of local search \cite{burke2014} are so effective,
and deep theoretical studies have been published 
on the performance of algorithms on specific classes of problems \cite{michiels2007}.

Our focus is on challenging problems for which it is hard to find optimal (or just ``good'') solutions. In section \ref{cardinality-monotonic} it will also be shown that all the example hard problems (classed as {\em PLS-Complete}) in \cite{michiels2007} have this same property that solutions thin out towards the optimum.
We call such functions ``cardinality-monotonic'' functions, and they are defined formally in Definition~\ref{def-cardinality-monotonic}.
The class of cardinality-monotonic functions is broad enough that the ``No Free Lunch'' theorems \cite{wolpert1997no}
hold for functions in this class.
However we will show that extra information available can make neighbourhood search effective
for this class of functions.
This information is enough to escape the conditions of the no free lunch theorem.

Key to the success of local search is the concept of {\em neighbourhoods},
and it is a particular property of neighbourhoods
(termed ``Neighbour's Similar Fitness'' or {\em NSF})
that makes it possible for neighbourhood search to perform better than random search.
An NSF neighbourhood is one that tends to link solutions with similar fitness.

The current study includes a probabilistic analysis of local search.
Given a current solution we evaluate the probability of finding a better solution.
A local search would try a neighbour of the current solution,
while a blind search would try a solution at random.

Intuitively, if the current solution has above average fitness
and the neighbourhood has the NSF property
then the expected value of the neighbour is higher than
the expected value of the randomly selected solution.

However, if nothing more is known about the neighbourhood,
the probability of improving on the current solution is less clear.
Indeed if the problem is not cardinality-monotonic, then even if the neighbourhood has NSF,
picking a neighbour is no more likely to improve on the current solution's fitness
than picking a solution at random. We formalise the definitions of {\em cardinality-monotonic} and {\em NSF},
and prove in Theorem~\ref{theo1} that under these definitions, 
the probability of a neighbour improving on a current solution
is indeed higher that the probability of a random solution improving on it.

\section{No Free Lunch Theorems}

The No Free Lunch (NFL) theorems~\cite{wolpert1997no} state that no single algorithm outperforms random search (equivalently, systematic linear search) when applied over all possible fitness functions. The theorems hold if all fitness functions defined over the given finite input space are equally likely. If the space of candidate solutions is infinite, the natural extension of the no free lunch theorems do not hold~\cite{auger2010continuous}. 

Other formulations of the NFLs in the literature have different emphasis. Whitley~\cite{whitley2000functions} proves that on average, no algorithm is better than random enumeration in locating the global optimum. Whitley~\cite{whitley2008focused} show that for all possible metrics, no search algorithm is better than another when its performance is averaged over all possible discrete functions. Serafino~\cite{serafino2013no} states that with no prior knowledge about the function, where any functional form is uniformly admissible, the information provided by the value of the function in some points in the domain will not say anything about the value of the function in other regions of its domain. 

More recent NFL variants assume special properties for the set of functions, their distribution, or their relationship with the algorithms. For example, the Sharpened No Free Lunch theorem shows that the result of the NFL holds even when we restrict consideration to certain subsets of function, such as any subset of functions closed under permutation~\cite{schumacher2001no,igel2005no}. Indeed we will make use of this result in section \ref{permutation} below.

However, if the set of possible fitness functions is restricted, then the conditions for the no free lunch theorems do not always hold. This does not invalidate NFL results, but caution against misapplication. 
For example Droste~\cite{Droste99} presents a restriction on function complexity, and shows on a small search space how this restriction on the class of functions enables the prediction of the no free lunch theorem to be violated: specifically local search algorithms can outperform a random generate-and-test. This extension is known as Almost No Free Lunch.

Christensen et al.~\cite{christensen2001can} characterise how effective optimisation can be under reasonable restrictions, and later generalised in \cite{whitley2006subthreshold}. The authors operationally define a method for answering the question of what makes a function searchable in practice, which involves defining a scalar field over the space of all functions. The method builds up information about the function by sampling, which is then used to guide the search. This algorithm can be expected to perform well if previous performance is an accurate indicator of future performance. There is, of course, no guarantee that this is the case.

Different from previous work, we introduce an abstract model of a problem and its landscape. The model does not distinguish between solutions of the same fitness. This level of abstraction frees us from concerns about specific local search algorithms, and enables us to address general mathematical properties.

\section{Context and Definitions}

In the following formalisation we represent a problem $(S,f)$ as a finite search space $S$ with a finite range of fitness values $\{f(S): s \in S\} \subseteq V = \{v_{min} \ldots v_{max}\}$. We define $ct_v$ to be the number of solutions in the search space with fitness value $v$. Without loss of generality we consider higher fitness values as better, in this formalisation. The search space size is
$|S| = \sum\limits_{v \in V} ct_v.$ The proportion of solutions $s$ with $f(s)=v$ and the proportion of solutions $s$ with values better than $v$ are
\begin{equation}
p_{v} = \frac{ct_v}{|S|}, \quad p^+_v = \frac{\sum_{v'>v} ct_{v'}}{|S|}.
\end{equation}

We now consider the landscape $(S,f,N)$ associated with the problem $(S,f)$. $N(s)$ returns the set of neighbours of solution $s$.  We will refer to the set of neighbours $\rm{Nf}(v)$ of solutions with fitness $v$:
\begin{equation}
\rm{Nf}(v) = \bigcup_{s \in S: f(s)=v} N(s)
\end{equation}
Finally, the proportion of such neighbours with fitness better than $v$ is:

\begin{equation}
pn^+_v = \frac {|\{s \in \rm{Nf}(v) : f(s)>v\}|} {|\rm{Nf}(v)|}.
\end{equation}

\section{Effective Neighbourhood Search}

neighbourhood search is effective at a current solution with fitness $v$ if the probability that a neighbour is fitter than $v$ is higher than the probability that an arbitrary solution is fitter than $v$.

\begin{definition}\label{effective}
Neighbourhood search is {\em effective} on a problem at a fitness $v$  in a landscape $(S,f,N)$ if $pn^+_v > p^+_v$
\end{definition}

This can never hold for all fitness values if the neighbourhood relation is symmetric. In such a case the probabilities across all solutions must balance so that the total probability of improvement is the same as the total probability of deterioration. However in many landscapes, there is a minimum value above which it holds.

\label{vge-improvement}
We say that neighbourhood search is effective in a landscape with global maximum fitness $v_{\text{max}}$, if there is a ``good enough'' fitness value $v_{ge} < v_{\text{max}}$ such that neighbourhood search is effective for all fitness values $v: v_{ge} \le v < v_{\text{max}}$. Naturally $v_{ge}$ should be sufficiently far from the optimum that it is relatively easy to find solutions with this fitness by random search.

This is formalised in Definition~\ref{LSeffective}:

\begin{definition}\label{LSeffective}
Neighbourhood search is \emph{effective in a landscape} $(S,f,N)$ if there is a ``good enough'' fitness value 
$v_{ge} < v_{\text{max}}$, where $v_{\text{max}}$ is the global optimum, such that $\forall v : v_{\text{max}} > v \ge v_{ge} \Rightarrow pn^+_v > p^+_v$ 
\end{definition}

Note first that neighbourhood search has a higher probability of improving than random search for all fitness values better than $v_{ge}$.
Consequently, if neighbourhood search is effective for the incumbent fitness, and a neighbour with better fitness is found, then neighbourhood search must also be effective for the neighbour's fitness.

Contrast this with random search. For any fitness value $v$, the probability $p^+_v$ that random search yields a solution better than $v$ is fixed.
Nothing can be learnt from previous random choices to improve this probability.

\section{Cardinality-Monotonicity}
\label{cardinality-monotonic}

The cardinality-monotonic property holds if the number of solutions $ct_v$ with given fitness $v$ decreases towards the optimum fitness.

\subsection{Typical Problems have fewer and fewer high quality solutions}

Many fitness functions are expressed as the sum of a set of terms, for example the Schwefel function:
\[
f(\textbf{x}) = f(x_1, x_2, ..., x_n) = 418.9829d -{\sum_{i=1}^{n} x_i \sin(\sqrt{|x_i|})}
\]
The number of terms in the sum increases with the number of variables.
Problems with a finite search space and fitness function expressed as a sum of this kind include the travelling salesman problem (TSP), quadratic assignment, and indeed all the examples of PLS-complete problems in Appendix C of \cite{michiels2007}. The PLS-complete problems are the hardest local search problems $(S,f)$ for which $f$ can be evaluated in polynomial time.

Problems $(S,f)$ whose functions expressed as a sum tend to have the cardinality-monotonic property.
If each term $t$ has a range of possible values $\min_t \ldots \max_t$, then the maximum sum $\sum_t \max_t$ can only be reached if every term takes its maximum value.  Similar only the sum of all the minima $\sum_t \min_t$ can lead to the minimum fitness value.  However for fitness value between these extreme there are many combinations of values that can form the same sum.
For example given 5 terms $t$ each of which can take values in the range $1 \dots 5$, the number of ways of creating each sum is:

\begin{figure}[!h]
    \centering
    \includegraphics[width=\linewidth]{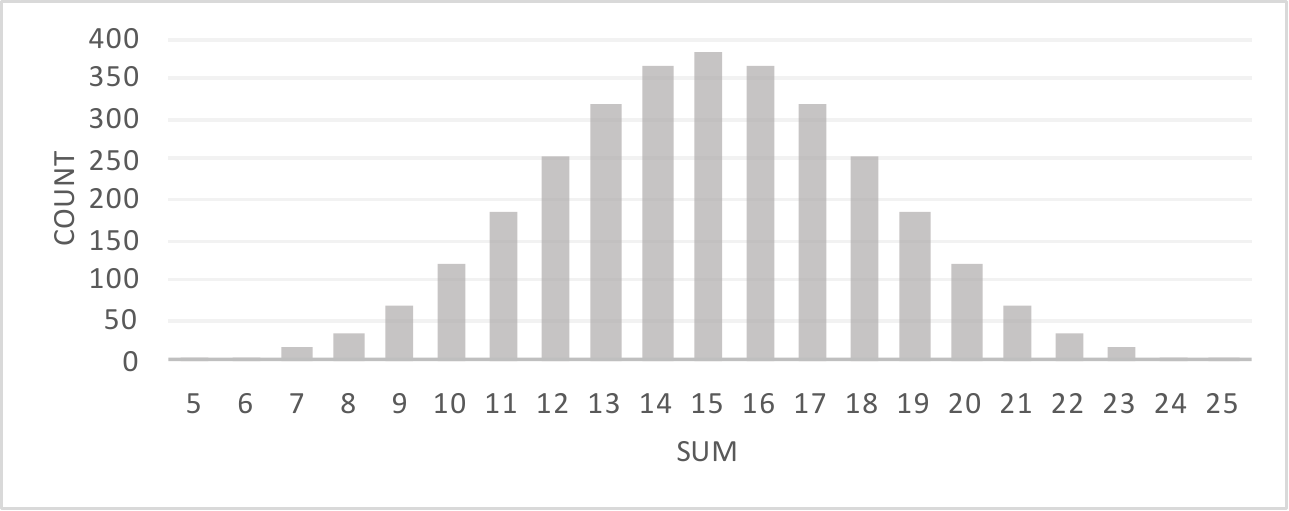}
    \caption{Total counts for 5 terms taking values in $1 \ldots 5$}
    \label{fig:ex}
\end{figure}

Consider the TSP, whose fitness function is the sum, over all the cities, of the distances to their successors in a solution.
If $c_j$ is the $j^{th}$ city visited on a route, and $d(a,b)$ is the distance between the cities $a$ and $b$, the cost of the route is
$\sum_j(d(c_j,c_{j+1}))$ 
An {\em assignment} for the TSP is an arbitrary assignment of a successor for each city, where the constraint that the tour must be a cycle is ignored.
The set of assignments for 5-city TSP, where each city had 5 possible successors at distances $1 \ldots 5$, has exactly the above distribution.  

Naturally in a solution to a TSP, the assignment must satisfy the constraint that the successors form a cycle. 
We call an assignment that satisfies the constraints of a problem a {\em feasible} assignment.
The cycle constraint does not mention the distance between any pair of cities, so there is no link between the fitness of an assignment in a TSP  and its feasibility.
Therefore, the restriction to feasible assignments for the TSP does not change the pattern of a decreasing number of solutions towards the optimum.

In a TSP with 12 cities, $1 \ldots 12$, where each city has 11 possible successors, we assigned a distance in the range $1 \ldots 20$ to each successor arbitrarily
\footnote{We created symmetric TSPs by giving a distance to each of the city-city pairs.  The process applies for any number of cities and distance values, but we describe it here for 12 cities and 20 distances. The 66 pairs take 20 different values (the 6 lowest distances occurring 4 times and the other distances thrice).  Numbering the cities $1$ to $12$, the pairs of cities are put in the order $<1,2>,<1,3>, \ldots <1,12>,<2,3>, \ldots <2,12>, \ldots <11,12>$ and assigned increasing distances}
and counted the number of feasible assignments for each route length, which is the fitness value for a TSP.
The minimum feasible value was 102, and the maximum 140, giving the following histogram:

\begin{figure}[!h]
    \centering
    \includegraphics[width=\linewidth]{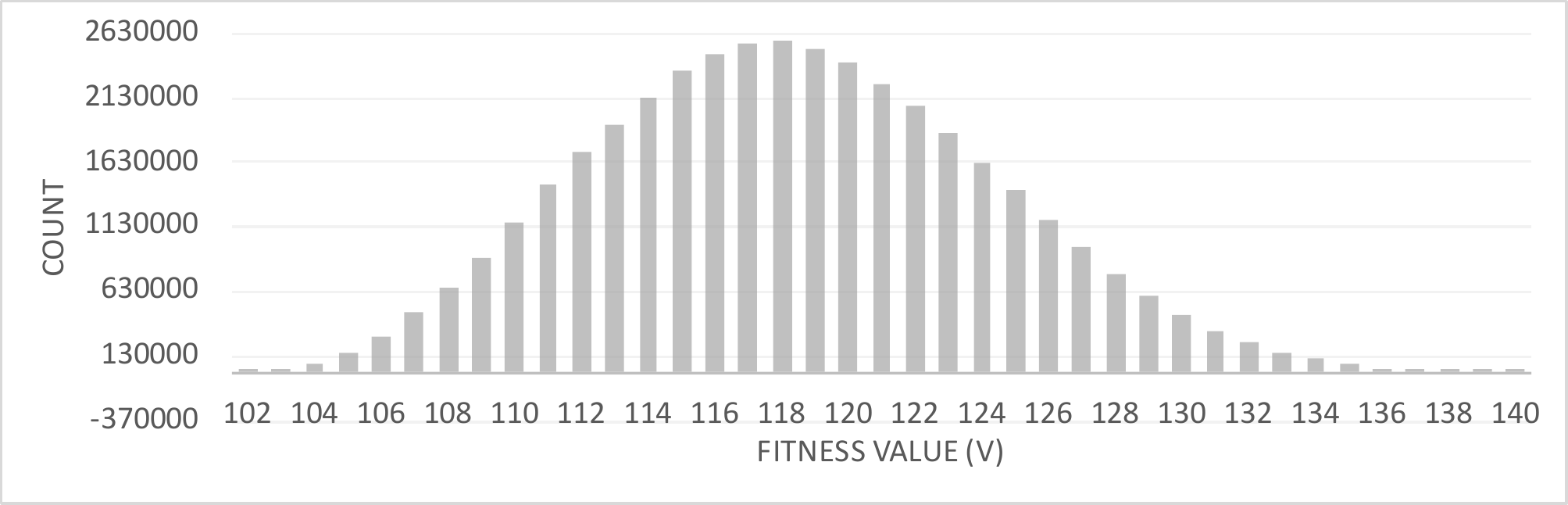}
    \caption{Counts of 12-city TSP solutions}
    \label{fig:tsp}
\end{figure}


Again there are relatively few optimal solutions, increasing to a modal (most common) fitness value of 118, and then decreasing again towards the worse fitness value of 140.

With this background, we can now define a problem as being {\em cardinality-monotonic} when the number of solutions $ct_v$ for each fitness value $v$ decreases monotonically from the modal fitness value to the optimum.  In case there are multiple modal fitness values, we choose the modal fitness value with the best fitness.

\begin{definition}\label{monotonicity}\label{def-cardinality-monotonic}
A search space and fitness function have the property of \emph{cardinality-monotonicity} if, above the highest modal fitness value $v_{mode}$, $ct_v$ decreases monotonically with increasing $v$: 
$$\forall v,v' \in V: v' \geq v \geq v_{\text{mode}} \rightarrow ct_{v'} \leq ct_v$$
\end{definition}

Note that the definition says nothing about poor fitness values below the modal fitness.  Often there is a decrease towards the worst fitness values, but the distribution of poor fitness values does not affect our proof of the effectiveness of neighbourhood search.

For higher fitness values, this definition indeed reflects the distribution of solutions in typical local search benchmarks, PLS-complete problems, and hard search and optimisation problems more generally.
The reason is that for PLS-complete problems, there is no link or correlation between the fitness of an assignment and its feasibility.
An exception to this occurs during branch and bound, and similar optimisation algorithms where constraints are added during search to exclude poor quality assignments. Nevertheless even in these cases, while the search space is dramatically pruned, the remaining `high quality'' solutions, still maintain the cardinality-monotonic property.

Many local search approaches do not exclude assignments which violate constraints, but rather recode the constraints as penalty functions and admit all assignments as candidate solutions, incurring additional penalties. In this case there is no infeasibility, but the resulting problem is likely to remain cardinality-monotonic for essentially the same reason - that fitness and feasibility are independent.
The new fitness function is the sum of the {\em underlying} fitness and all the {\em penalties}.
Naturally a penalty cannot improve the fitness of a solution.
Solutions with all underlying fitness values are equally likely to incur penalties.  
Thus if there are monotonically decreasing values of $ct_v$ for high quality underlying fitness values $v$, then the counts of the new fitness values will still, with high probability, be monotonically decreasing.

The proof below relies on strict cardinality-monotonicity.
We recognise, however, that there are problems which are nearly cardinality-monotonic, but might have, for example, no solutions at a certain good fitness value, though there are solutions with better fitness values.
The proof can be extended to handle problems which are not strictly cardinality-monotonic, but how far a problem can diverge from strict cardinality-monotonicity and still support a proof that neighbourhood search is effective, is an open question.
In this paper we therefore tackle only the strict case.

\subsection{No Free Lunch for Cardinality-Monotonic Problems}
\label{permutation}
The no free lunch theorem goes through even if we restrict the problem class to be cardinality-monotonic. Firstly we show that the class of cardinality-monotonic problems is closed under permutation \cite{Joyce2018}
\begin{definition}
Let G be a set of functions mapping a finite domain $X$ to a finite range $Y$.
We say G is closed under permutation iff for any permutation $\phi: X \rightarrow X$, we have $f \in G \Rightarrow f_{\phi} \in G$.
\end{definition}
\begin{proof}
Consider a fitness value $y \in Y$. In the problem $(S,f)$, 
if $f(s)=y$, then in $(S,f_{\phi})$, $f_{\phi}(\phi^{-1}(s)) = y$.
$ct_y$ is the number of values (solutions) $s \in S$ for which $f(s)=y$ and is also the number of values $\phi^{-1}(s) \in S$ for which $f_{\phi}(\phi^{-1}(s)) = y$.
This means that in the problem $(S,f_{\phi})$, $ct_y$ remains unchanged.
Therefore, $(S,f)$ is cardinality-monotonic if and only if $(S,f_{\phi})$ is.
$\quad \blacksquare$
\end{proof}

The sharpened no free lunch theorem \cite{schumacher2001no,igel2005no} states that the no free lunch theorem holds over any set of functions closed under permutation.
It follows that the no free lunch theorem indeed holds for the class of cardinality-monotonic problems.

\section{Neighbours with Similar Fitness (NSF)}\label{sec:nsf}

By definition, PLS problems \cite{michiels2007} contain  a combinatorial optimization problem and a {\em reasonable} neighborhood function.
It is easy to invent "unreasonable" functions which make all combinatorial optimisation problems easy to solve with local search. We could simply insist that (a) every solution, except optimal ones, has at least one neighbour, and that (b) all the neighbours have better fitness than the solution.Then every hill climb will go straight to the global optimum!

An unreasonable aspect of this neighbourhood is the imbalance in each neighbourhood between better and worse neighbours. This is unreasonable because in order to construct neighbourhoods that are skewed towards better neighbours, it would be necessary to do enough to solve the problem beforehand. 

In this paper we assume neighbourhoods are reasonable in the sense that they are not skewed towards better (or worse) solutions: we term this "unskewed".
Consider neighbours of solutions with a fitness value of $v$. The number of their neighbours which differ from $v$ by $\delta$ is $ctn_{v,\delta}$:
\begin{equation}
ctn_{v,\delta} = |\{s' \in \rm{Nf}(v) : f(s') = v \pm \delta\}|
\end{equation}

In an unskewed neighbourhood, how many of these neighbours have better fitness $v + \delta$?
We define the number of neighbours of solutions with fitness $v$, with fitness greater by $\delta$ to be:
\begin{equation}
ctn^+_{v,\delta} = |\{s' \in \rm{Nf}(v) : f(s') = v + \delta\}|
\end{equation}
If the neighbourhoods are unskewed the proportion of better fitness neighbours must be the same as in the search space as a whole. The proportion of better neighbours is $pn^+_{v,\delta}$ and the proportion in the search space as a whole is $p^+_{v.\delta}$:
\begin{equation}
pn^+_{v,\delta} = \frac{ctn^+_{v,\delta}}{ctn_{v,\delta}}, \quad p^+_{v.\delta} = \frac{ct_{v+\delta}}{ct_{v \pm \delta}}
\end{equation}

\begin{definition}
In an {\em unskewed} neighbourhood $pn^+_{v,\delta} = p^+_{v.\delta}$
\end{definition}

Clearly if the fitness of the neighbours of a solution are independent of the solution's fitness, a local move will be no better (or worse) than random search.
The NSF property holds for a neighbourhood operator if, and only if, there is a correlation between the fitness of a solution and its neighbours.
Specifically the neighbours should have similar fitness values.

Having argued that skewed neighbourhoods are unreasonable because they are hard to construct, we need to show that NSF neighbourhoods are by contrast easy to construct. The very same property of typical fitness functions - that they can be expressed as a sum of terms - is useful for constructing simple NSF neighbourhoods. As a first example, let us return to the TSP, whose fitness function is a sum of distances.  The number of terms in the sum is the number of cities in the TSP.

However the 2-swap neighbourhood for symmetric TSPs, changes only two distances in the sum - no matter how many cities there are in the TSP.  This is, in fact,
the smallest change possible maintaining the constraint that the tour must be a cycle.  
By ensuring that $N-2$ distances remain the same (where $N$ is the number of cities), the 2-swap generates neighbours that have similar fitness.

For many problems the neighbourhood operator is defined so as to change only a few terms in the sum which expresses the fitness function.  
The maximum satisfiability problem is to find an assignment to truth variables that minimises the number of unsatisfied clauses.  A clause is just a disjunction of some truth variables (or none) and some negated truth variables (or none).
A neighbour for this problem is found by changing the truth value of one variable ("flipping" a variable).
If the clause length is restricted to just three (variables and negated variables), and if there are, say, 100 truth variables in the problem, then only 
$1-\frac{99^3}{100^3} = 0.03$ of the clauses are likely to contain any given variable.
Thus after flipping a variable 97\% of the clauses will remain unchanged.
Consequently the fitness of a neighbour found by flipping a variable is likely to be similar to the original fitness.

The same argument applies to most problems whose fitness function is a sum of terms in which the number of terms increases with the number of variables in the problem.
Any neighbourhood operator that changes the value of a single variable, or a small set of variables, will only change the value of a small fraction of the terms in the sum.
Consequently neighbours are likely to have similar fitness. The proportion of solutions that differ in fitness by $\delta$ from a given fitness $v$ and the proportion of neighbours (solutions in $Nf(v)$) that differ from $v$ by $\delta$, are calculated as
\begin{equation}
p_{v,\delta} = \frac{ct_{v,\delta}}{|S|}, \quad pn_{v,\delta} = \frac{ctn_{v,\delta}}{|Nv(v)|}
\end{equation}

We use the expressions $p_{v,\delta}$ and $pn_{v,\delta}$ in the following definition of the Neighbours with Similar Fitness (NSF) property. The property states that neighbours tend to have similar fitness than non-neighbours. We denote the set $\{\delta: 0<\delta \leq v_{\text{max}}-v_{\text{ge}}\}$ by the symbol $\Delta$.

\begin{definition}\label{nsf}
A landscape has the \emph{Neighbours with Similar Fitness} (NSF) property if across all fitness values $v$ the value of the expression
$ pn_{v,\delta} - p_{v,\delta}$
decreases monotonically with increasing $\delta \in \Delta$,
$pn_{v,1}>p_{v,1}$ and
$\sum_{\delta \in \Delta} pn_{v,\delta} - p_{v,\delta} \geq 0$
\end{definition}

This property means that for each fitness value $v$
\begin{compactitem}
\item
the probability that a neighbour of $s$ has fitness $f(s)$ close to $v$ is higher than the probability of a solution in the search space as a whole having fitness close to $v$,
\item
as the fitness value difference $\delta \in \Delta$ grows, the difference between the probability $pn_{v,\delta}$ of a neighbour with fitness differing by $\delta$ from $v$, and the probability $p_{v,\delta}$ of a solution in the whole search space with fitness differing by $\delta$ from $v$, monotonically decreases.
\item 
The increased probability of neighbours with a small difference in fitness outweighs any decreased probability of other fitness differences outside the range $1 \ldots v_{\rm{max}}-v_{\rm{ge}}$
\end{compactitem}


The NSF property is a special case of the high-locality property, defined as ''a small change to the genotype should on average result in a small change to the phenotype and fitness value.''. However, the high-locality property does not impose the conditions presented in Definition~\ref{nsf}.

\subsection{Good enough fitness}
Neighbourhood search was defined above to be effective in a landscape if there is a good enough fitness value $v_{ge}$ at and above which it is effective at everywhere.

This fitness in a cardinality-monotone problem is halfway between the optimum and the modal fitness level.  The number of solutions at each fitness level monotonically increases from the optimum to this modal fitness, and the good enough fitness value $v_{ge}$ is at the midpoint.
\begin{definition}\label{eq:ge}
The good enough fitness level $v_{ge}$ is defined to be
$(v_{\text{max}}-v_{\text{mode}})/2$
\end{definition}

In general there is a good proportion of solutions with fitness as good as or better than $v_{ge}$.
In the first distribution in fig \ref{fig:ex}, the proportion above $v_{ge}$ is 7\%, so random search would find such a value after about 14 trials on average.
In the second distribution in table \ref{fig:tsp}, the proportion is 6.3\% so random search would find such a value in less than 16 trials on average.
At this level and above, cardinality-monotonic problems have a useful property.
Take any fitness value $v > v_{ge}$. For all values of $0< \delta <  v_{\text{max}}-v$ the ratio $p^+_{v,\delta} = \frac{ct_{v+\delta}}{ct_{v \pm \delta}}$ decreases with $\delta$.  
This means that for solutions with similar fitness to $v$, the proportion of improving solutions is higher than for solutions with fitness further from $v$.

\section{Conditions for Local Improvement to be Effective}\label{sec:examples}

Assuming that the cardinality-monotonicity and NSF conditions are satisfied in a landscape, we prove in \ref{apptheo1} that, under definition \ref{LSeffective}, neighbourhood search is effective.

\begin{figure}[!ht]
\centering
\includegraphics[width=0.7\linewidth]{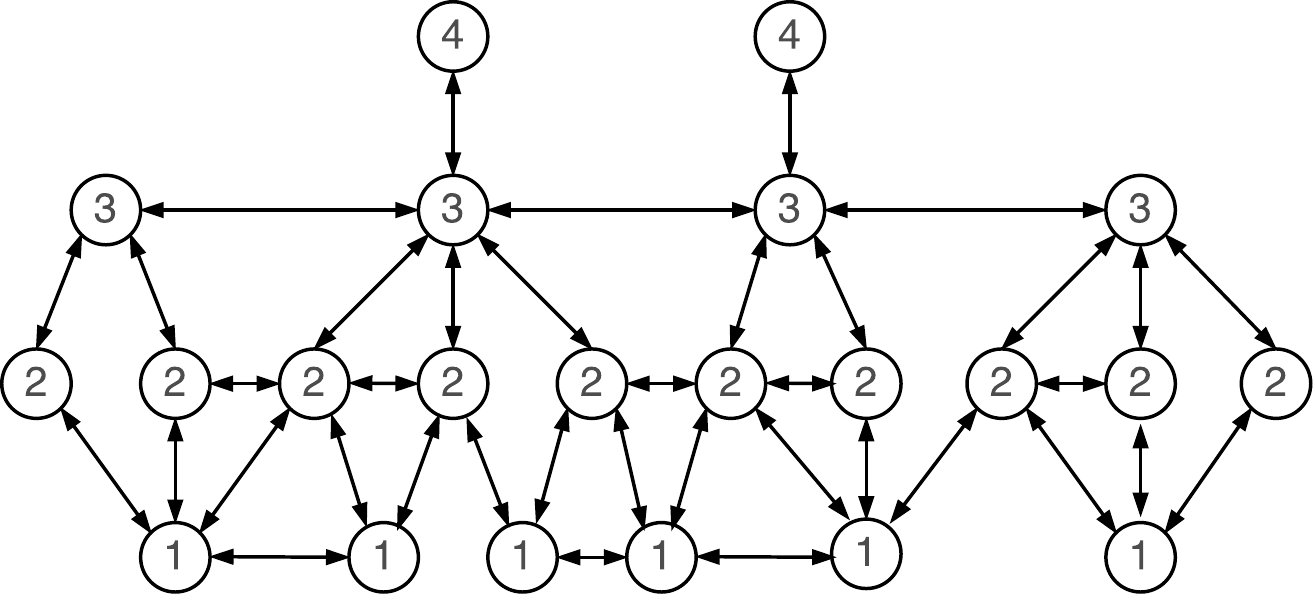}
\caption{A simple landscape satisfying both properties: NSF and cardinality-monotonicity. Nodes represent solutions, labels represent fitness values, and arrows represent neighbour relationships.}\label{nsfla}
\end{figure}

Figure~\ref{nsfla} illustrates a landscape satisfying both properties -- NSF and cardinality-monotonicity. There are 4 fitness values, 1..4, and the number of solutions for each fitness value decreases monotonically above the modal fitness value of 2 (10 solutions with fitness 2, 4 with fitness 3, and 2 solutions with fitness 4). The landscape satisfies NSF since all neighbours of each solution have fitness value differing by 0, +1 or -1; there are few neighbours with the same fitness; and the proportion of neighbours differing in fitness by +1 and -1 is the same as the proportion in the whole solution space.

In this landscape the probability $p^+_3$ of randomly picking a solution with value 4 is $2/22 = 0.091$.  However starting from a solution with value 3, the probability $pn^+_3$ of selecting a neighbour with value 4 is $2/15 = 0.13$.
Thus in this example neighbourhood search {\em is} effective. 

\section{Proof that cardinality-monotonicity and NSF Justify neighbourhood search}
\label{apptheo1}
\begin{theorem}\label{theo1}
If the landscape has cardinality-monotonicity and NSF, then
$$\forall v > v_{ge} : p^+_v < pn^+_v$$
\end{theorem}
The theorem states that the probability of a neighbour of a solution with fitness $v$ having fitness $> v$ is greater than the probability of an arbitrary node with fitness $> v$. To support this proof, we first show that for good enough values of $v$,  $p^+_{v,\delta}$ decreases with $\delta$.
This result is independent of the landscape, and depends only on the cardinality-monotonicity condition.
\begin{lemma}\label{lemma}
$\forall v > v_{ge}, \delta \in \Delta: p^+_{v,\delta}$ is monotonically decreasing 
\end{lemma} 


Recall that $\Delta = \{\delta: 0<\delta \leq v_{\text{max}}-v_{\text{ge}}\}$.
\begin{proof}\label{proof:lemma}
\textbf{Lemma \ref{lemma}}
\end{proof}
We consider two cases:
$\delta < v_{\text{max}}-v$ and $\delta = v_{\text{max}}-v$

\textbf{Case1:} $\delta < v_{\text{max}}-v$. By the cardinality-monotonicity assumption
\begin{align*}
&v_{ge}<v<v_{\rm{max}} \Rightarrow ct_{v+\delta+1} = ct_{v+\delta} \cdot \alpha, \text{ for some } \alpha \leq 1\\
&v>v_{ge} \wedge \delta < (v_{max}-v) \Rightarrow (v-\delta) > v_{\text{mode}}\\
&(v-\delta) > v_{\text{mode}} \Rightarrow ct_{(v-\delta)-1} = ct_{v-\delta} \cdot \beta, \text{ for some } \beta \geq 1.
\end{align*}

Therefore
\begin{align*}
p^+_{v,\delta+1} = \frac{ct_{v+\delta+1}}{ct_{v+\delta+1} + ct_{(v-\delta)-1}}= \frac{\alpha \cdot ct_{v+\delta}}{\alpha \cdot ct_{v+\delta} + \beta \cdot ct_{v-\delta}}.
\end{align*}
Since $\alpha \leq \beta$, then
\begin{align*}
p^+_{v,\delta+1} &\leq \frac{\alpha \cdot ct_{v+\delta}}{\alpha \cdot ct_{v+\delta} + \alpha \cdot ct_{v-\delta}}\leq \frac{ct_{v+\delta}}{ct_{v+\delta} + ct_{v-\delta}}\leq p^+_{v,\delta}
\end{align*}

\textbf{Case 2:} $\delta = v_{\text{max}}-v$. In this case $ct_{v+\delta+1} = 0$, and therefore $p^+_{v,\delta+1} = 0$. Since $ct_{v_{\text{max}}}>0$, it follows that $p^+_{v,\delta} > 0$. Therefore $p^+_{v,\delta} > p^+_{v,\delta+1}$. 

\noindent Combining both cases, we have 
$\forall \delta \geq v_{\text{max}}-v: p^+_{v,\delta}$ 
is monotonically decreasing. $\quad \blacksquare$

\begin{proof}~\label{proof:th1}
\textbf{Theorem \ref{theo1}}

\emph{We prove} $$v > v_{ge} \Rightarrow  pn^+_v - p^+_v > 0 $$
\end{proof}

Since the neighbourhood is unskewed:
\begin{align*} 
pn^+_v& =\sum\limits_{\delta \in \Delta} pn_{v,\delta} \cdot pn^+_{v,\delta}=\sum\limits_{\delta \in \Delta} pn_{v,\delta} \cdot p^+_{v,\delta}.
\end{align*}

Also, $p^+_v = \sum\limits_{\delta \in \Delta} p_{v,\delta} \cdot p^+_{v,\delta}.$

Since $\delta > v_{\rm{max}}-v \Rightarrow pn^+_v = p^+_v = 0$, we can rewrite the expression $pn^+_v - p^+_v$ as follows:
\begin{align*}
pn^+_v - p^+_v&=  \sum\limits_{\delta \in \Delta} pn_{v,\delta} \cdot p^+_{v,\delta} - \sum\limits_{\delta \in \Delta} p_{v,\delta} \cdot p^+_{v,\delta} \\
&= \sum\limits_{\delta \in \Delta} (pn_{v,\delta}-p_{v,\delta}) \cdot p^+_{v,\delta}.
\end{align*}

By NSF, $$\sum\limits_{\delta \in \Delta} pn_{v,\delta} \geq \sum\limits_{\delta \in \Delta} p_{v,\delta}.$$

Also $pn_{v,1}>p_{v,1}$ and $pn_{v,\delta}-p_{v,\delta}$ is monotonically decreasing with $\delta$, so there must be some $x \in \Delta$ for which
\begin{equation}\label{eq:strictness}
\begin{split}
\forall \delta \leq x.  & pn_{v,\delta}-p_{v,\delta} > 0 \\
\forall \delta > x. & pn_{v,\delta}-p_{v,\delta} \leq 0  
\end{split}
\end{equation}

By lemma \ref{lemma} 
\begin{equation}
\begin{split}
\forall \delta \leq x. p^+_{v,\delta} \geq p^+_{v,x}\\
\forall \delta > x. p^+_{v,\delta} \leq p^+_{v,x+1}
\end{split}
\end{equation}

Consequently:
\begin{equation*}
\begin{split}
&\sum\limits_{\delta \in \Delta} (pn_{v,\delta}-p_{v,\delta}) \cdot p^+_{v,\delta} \\
=& \sum\limits_{\delta \leq x \in \Delta} (pn_{v,\delta}-p_{v,\delta}) \cdot p^+_{v,\delta} +
\sum\limits_{\delta >x \in \Delta} (pn_{v,\delta}-p_{v,\delta}) \cdot p^+_{v,\delta} \\
=& \sum\limits_{\delta \leq x \in \Delta} (pn_{v,\delta}-p_{v,\delta}) \cdot p^+_{v,\delta} -
\sum\limits_{\delta >x \in \Delta} (p_{v,\delta}-pn_{v,\delta}) \cdot p^+_{v,\delta} \\
>& \sum\limits_{\delta \leq x \in \Delta} (pn_{v,\delta}-p_{v,\delta}) \cdot p^+_{v,x} -
\sum\limits_{\delta >x \in \Delta} (p_{v,\delta}-pn_{v,\delta}) \cdot p^+_{v,x+1}
\end{split}
\end{equation*}

For ease of readability, we introduce the following notations which will be helpful in the proof:
\begin{align*}
&pn_{v,\leq x} = \sum\limits_{\delta \leq x} pn_{v,\delta} \ \ \ & pn_{v, >x} = \sum\limits_{\delta > x} pn_{v,\delta}\\
&p_{v,\leq x} = \sum\limits_{\delta \leq x} p_{v,\delta} \ \ \ & p_{v, >x} = \sum\limits_{\delta > x} p_{v,\delta}
\end{align*}

Due to NSF 
$$\sum\limits_{\delta \in \Delta} pn_{v,\delta} \geq \sum\limits_{\delta \in \Delta} p_{v,\delta}.$$

Thus

\begin{equation*}
\begin{split}
& (pn_{v,\leq x} + pn_{v, >x}) > ( p_{v,\leq x} +  p_{v, >x})\\
\therefore & (pn_{v,\leq x} - p_{v,\leq x}) + (pn_{v, >x} - p_{v, >x}) > 0\\
\therefore & (pn_{v,\leq x} - p_{v,\leq x}) - (p_{v, >x} - pn_{v, >x}) > 0\\
\therefore & (pn_{v,\leq x} - p_{v,\leq x}) > (p_{v, >x} - pn_{v, >x})
\end{split}
\end{equation*}

We conclude as required:
\begin{equation*}
\begin{split}
& pn^+_v - p^+_v\\
>& \sum\limits_{\delta \leq x \in \Delta} (pn_{v,\delta}-p_{v,\delta}) \cdot p^+_{v,x} -
\sum\limits_{\delta >x \in \Delta} (p_{v,\delta}-pn_{v,\delta}) \cdot p^+_{v,x+1}\\
=& \quad (pn_{v,\leq x} - p_{v, \leq x})\cdot p^+_{v,x} - (p_{v, >x} - pn_{v, >x})\cdot p^+_{v,x+1}\\
>& \quad (pn_{v,\leq x} - p_{v, \leq x})\cdot p^+_{v,x} - (p_{v, >x} - pn_{v, >x})\cdot p^+_{v,x}\\
=& \quad ((pn_{v,\leq x} - p_{v, \leq x})- (p_{v, >x} - pn_{v, >x}))\cdot p^+_{v,x}\\
> & \quad 0 \quad \blacksquare
\end{split}
\end{equation*}

\section{Conclusion}
\label{conclusion}
This paper identifies minimal conditions under which local search is expected to outperform random search. The key condition is quite intuitive: neighbouring solutions should have similar fitness values. However, a further property, ``cardinality-monotonicity'' is needed for local search to outperform random search. A search space has \emph{cardinality-monotonicity} property if starting from the fitness value which occurs most often, better and better fitness values occur less and less often in the search space. This condition seems reasonable in that it holds - broadly if not always strictly - in typical combinatorial problems. In \emph{all} reasonable landscapes (not skewed towards better-, or worse-, fitness neighbours) which have these properties, given a solution with good enough fitness (as defined in the paper), a hill climb starting from that solution is more likely to improve than a random search.

Previous work on problems suitable for local search have either proposed certain fitness functions, e.g. \cite{Droste99}, or investigated landscapes within which random walks have certain properties, e.g.  \cite{Angel98}. The properties introduced in this paper are both more general than those analysed in previous work, and shown to hold in typical hard local search problems. Our model of a problem and its landscape is abstract. Solutions of the same fitness are not distinguished, and the only information about the landscape is the number of neighbours of any given fitness that have each other fitness value. This level of abstraction frees us from concerns about specific local search algorithms, and enables us to address general mathematical properties.

\bibliographystyle{named}

\end{document}